\documentclass{article}
\usepackage{spconf,amsmath,graphicx}
\usepackage{xcolor}
\usepackage{amssymb,amsmath,epsfig,cite,url,multicol,multirow}
\usepackage{hyperref}
\usepackage{graphicx}
\usepackage{float}
\usepackage{comment}
\usepackage{float}
\usepackage{diagbox}
\floatstyle{plaintop}
\restylefloat{table}

\title{Leveraging Acoustic and Linguistic Embeddings from Pretrained speech and language Models for Intent Classification}
\name{Bidisha Sharma, Maulik Madhavi and Haizhou Li\thanks{This research is supported by the Agency for Science, Technology and Research (A*STAR) under its AME Programmatic Funding Scheme (Project No. A18A2b0046).}}
\address{Department of Electrical and Computer Engineering,\\
  National University of Singapore, Singapore}
\begin{document}
%
\maketitle
\begin{abstract}
Intent classification is a task in spoken language understanding. An intent classification system is usually implemented as a pipeline process, with a speech recognition module followed by text processing that classifies the intents. There are also studies of end-to-end system that take acoustic features as input and classifies the intents directly. Such systems don't take advantage of relevant linguistic information, and suffer from limited training data. In this work, we propose a novel intent classification framework that employs acoustic features extracted from a pretrained speech recognition system and linguistic features learned from a pretrained language model. We use knowledge distillation technique to map the acoustic embeddings towards linguistic embeddings. We perform fusion of both acoustic and linguistic embeddings through cross-attention approach to classify intents. With the proposed method, we achieve 90.86\% and 99.07\% accuracy on ATIS and Fluent speech corpus, respectively.
\end{abstract}
\vspace{-0.3cm}
\section{Introduction}
\label{sec:intro}

With the increasing adoption of voice-operated interfaces in smart devices, one of the research topics in spoken language understanding (SLU) is to achieve more natural, intuitive, robust and effective interaction~\cite{tur2011spoken,bhargava2013easy,ravuri2015recurrent,sarikaya2014application}. Intent classification refers to inferring the {\it meaning} or {\it intention} of a spoken utterance, which is a crucial component of SLU~\cite{bhargava2013easy}. The voice interfaces decide how to respond according to the perceived intent of a particular spoken utterance from the user.  

The typical SLU system architecture follows a pipeline approach, that consists of two components, first, an automatic speech recognition (ASR) system decodes the input speech into text transcription, that is followed by a natural language understanding (NLU) module to classify the intent from ASR output text~\cite{tur2011spoken}. 

The pipeline approach has a few limitations. First, pipeline components are optimized separately under different criteria. The ASR module is optimized to minimize the word error rate (WER), while the NLU module is typically trained on the clean text (original transcription). The ASR performance varies largely depending on the noisy environmental conditions resulting in erroneous transcription, which subsequently affects the performance of the NLU module. As a result, there is an obvious mismatch between training and testing conditions, which may limit the performance. Second, all words are not equally important for the intent classification task. Some words carry more weight towards the meaning of the utterance than others that is reflected in the linguistic prosody. Unfortunately, the pipeline approach has not taken  such prosodic information into consideration.

The end-to-end approaches represent one of the solutions to the above issues arising from pipeline approaches~\cite{qian2017exploring,serdyuk2018towards}. They are widely adopted in ASR~\cite{amodei2016deep,chan2016listen,soltau2016neural}, speech synthesis~\cite{oord2016wavenet}, machine translation~\cite{sutskever2014sequence,bahdanau2014neural,gehring2017convolutional}. Inspired by their success, Serdyuk et al.~\cite{serdyuk2018towards} introduces an end-to-end modeling approach for intent classification, where the features extracted from the speech signal are directly mapped to intents or other SLU task, without intermediate ASR and NLU components.

Lugosch et al.~\cite{serdyuk2018towards} study an end-to-end encoder-decoder framework for speech-to-domain and intent classification. Other studies reveal that along with the speech features,  an intermediate text representation is also crucial for the quality of the predicted semantics. Haghani et al.~\cite{haghani2018audio} propose a joint model that predicts both words and semantics and achieves a good performance using a large train database.

To benefit from the phonetic recognition of speech, another study~\cite{lugosch2019speech} explores the use of an ASR model for intent classification. In particular, a pre-trained ASR model is used to extract acoustic features from speech signals. Similarly, to address the issue of limited data, different neural network architectures are introduced in~\cite{lugosch2019speech, renkens2018capsule, poncelet2020multitask, mitra2019leveraging}. Due to lack of large speech databases for SLU tasks, authors of ~\cite{lugosch2019using} propose to use synthesized speech for training end-to-end SLU systems. 

Transfer learning has been successfully applied in intent classification tasks to align the acoustic embeddings extracted from an ASR system to linguistic embeddings extracted from a language model, such as bidirectional encoder representations from transformers (BERT)~\cite{huang2020leveraging, denisov2020pretrained}. These approaches use linguistic representations derived from acoustic features using transfer learning to classify intents. We consider that the original acoustic features or embeddings also carry significant information for SLU tasks.


Human perception and interpretation of a spoken utterance relies on both prosodic and linguistic information intact in the speech signal. In this work, we aim to benefit from both acoustic and linguistic representations, and leverage information extracted from pre-trained acoustic and language models. In particular, the use of pre-trained models helps to overcome the issue of unavailability of a large speech database for targeted SLU tasks. The transfer learning approach assists us to derive linguistic representation from the acoustic embeddings of speech. 

Specifically, we derive the acoustic embeddings from a pre-trained ASR model, and learn to derive linguistic embeddings that are close to those from a pre-trained BERT model by applying transfer learning. We then combine both acoustic and linguistic embeddings into a unified representation through a cross-attention module. With cross attention, the attention mask for one modality is used to highlight the extracted features in another modality. The strategy of combining both acoustic and linguistic embeddings makes this work significantly different from the existing transfer learning based intent classification methods~\cite{huang2020leveraging, denisov2020pretrained}. We note that the proposed framework exploits information extracted from only speech signal and we do not use ASR output text. 


\vspace{-0.3cm}
\section{Acoustic-linguistic network with transfer learning}\label{ProposedApproach}
\begin{figure}[h]
 \centering
\centerline{\epsfig{figure=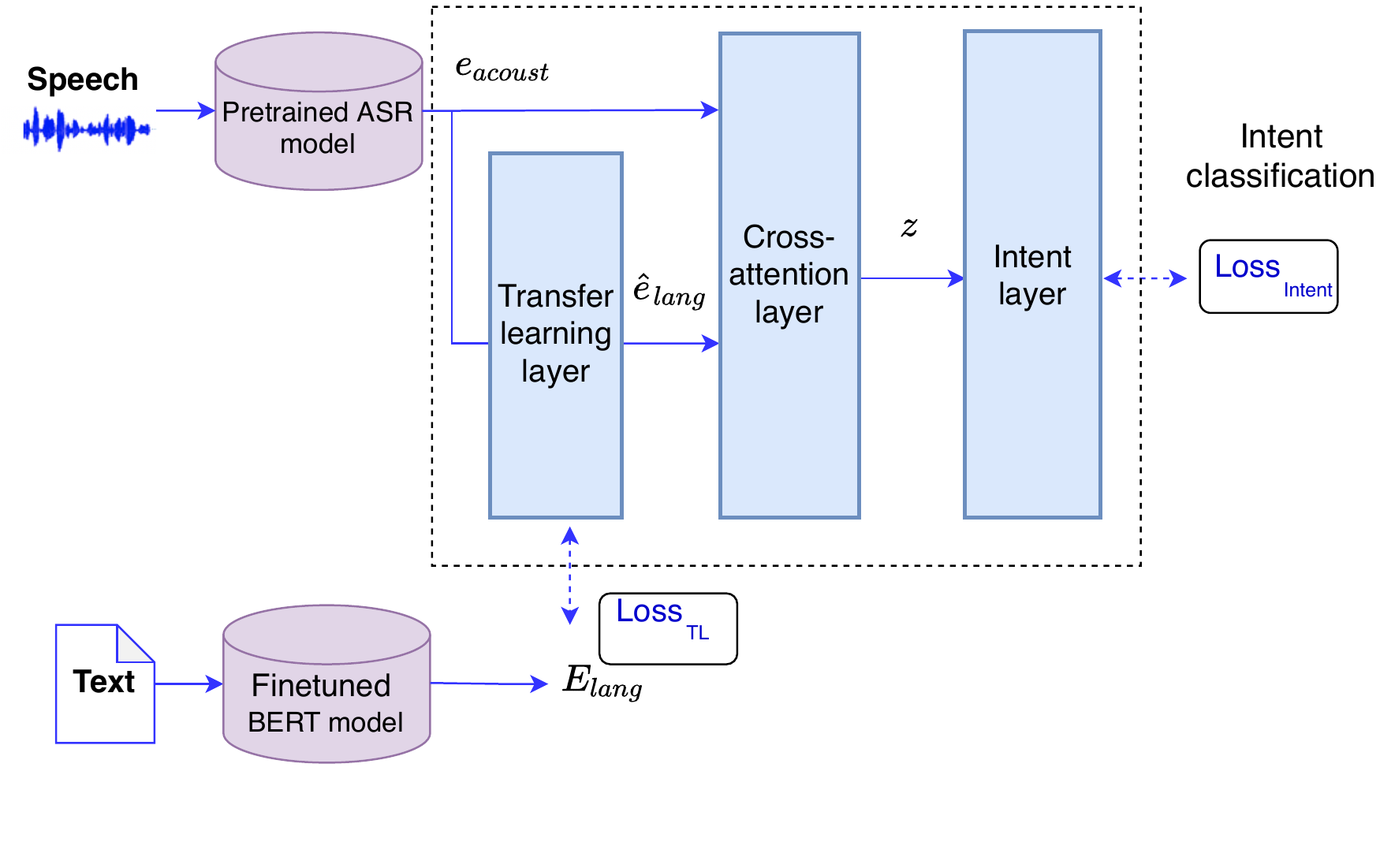,scale=0.51, trim=10 0 0 0 ,clip}}
\vspace{-.8cm}
{\ninept \caption{Block diagram of proposed acoustic-linguistic network (ALN) with transfer learning for intent classification, where the dotted block highlights our contribution.}
\vspace{-.4cm}
\label{fig:block-diagram}}
\end{figure}

\vspace{-0.2cm}
In Figure~\ref{fig:block-diagram}, we provide an overview of the proposed acoustic-linguistic network (ALN) with transfer learning. We leverage on the information extracted from two pretrained models, which are the ASR model and BERT model. Initially, we extract the  acoustic embeddings from the pretrained ASR model as described in~\cite{lugosch2019speech}. The latent space of these acoustic embeddings is transformed to NLU output space for the downstream task using the transfer learning layer. We refer to the derived embeddings as {\it ALN linguistic embeddings}. The acoustic embeddings extracted from ASR and ALN linguistic embeddings are effectively fused together for intent classification. The framework is optimized using two loss functions, which are transfer learning loss and intent loss. Each of these components are described in detail below. 

\vspace{-.4cm}

\subsection{Pre-trained ASR Model}\label{Pretrained_ASR_Model}
We follow the same strategy to use a pre-trained ASR model as described in~\cite{lugosch2019speech}\footnote{https://github.com/lorenlugosch/end-to-end-SLU}.
This model is a deep neural network consisting of a stack of modules. The first module takes as input the audio signal and outputs a sequence of hidden representations to represent the phonetic content of input speech.  
The second module takes the hidden phonetic representations as input and outputs 256-dimensional hidden word representation, which are used as acoustic embeddings ($e_{acoust}$) in this work. Further details of the pre-trained ASR model can be found in~\cite{lugosch2019speech}.  We freeze this pre-trained ASR model, and use it as it is for downstream intent classification task.
\vspace{-0.4cm}
\subsection{Fine-tuned BERT Model}
\label{sec:pretrained-nlu}
 BERT achieves state-of-the-art performance in sentence classification and other natural language processing tasks~\cite{devlin2019bert,sentencebert}. In order to incorporate domain-specific linguistic information, there is need to finetune the original BERT model. We fine-tune a pre-trained \verb|bert-base-uncased| model~\cite{devlin2019bert} for intent classification task. Inspired by \cite{huang2020leveraging}, we perform two-stage finetuning approach, first we fine-tune the BERT weights for masked language model (MLM) task with language model heads. We follow the masking procedure as recommended in \cite{devlin2019bert}. Next, using these weights, we further fine-tune BERT model with intent classification label as target. The experimental details are given in Section \ref{Experimentsetup}.

We apply the fine-tuned BERT model to capture linguistic information for intent classification. 
From sequence output of BERT  ($e_{lang}$), we use mean pooling and obtain the BERT embeddings $E_{lang}$, where $E_{lang}={MeanPool}(e_{lang})$. We further use this finetuned model as the teacher model in transfer learning stage.
 
\vspace{-0.4cm}
\subsection{Transfer Learning Layer}
Using the transfer learning layer, we aim to derive linguistic representation from the acoustic embeddings of speech that closely resemble the BERT embeddings. This is a linear layer that transforms the 256-dimensional acoustic embeddings to 768-dimensional linguistic embeddings. We follow teacher-student learning method as described in ~\cite{hinton2015distilling,li2017large} to learn the transfer learning layer acting as a student model. The BERT model as discussed in Section \ref{sec:pretrained-nlu} is considered as teacher model. We learn the transfer learning layer to produce embeddings that are closer to the BERT embeddings. 
We obtain ALN linguistic embeddings ($\hat{e}_{lang}$) as the output of the transfer learning layer, which is used in the next modules for intent classification. We note that the ALN linguistic embeddings are frame-aligned with the acoustic embeddings. Both sequences have the same length.  

We use mean square error (MSE) loss for backpropagation of this layer, which is defined as, 
\vspace{-0.2cm}
\begin{equation}
\vspace{-0.2cm}
Loss_{TL} = \text{MSE}(E_{lang}, \hat{E}_{lang}),
\vspace{-0.1cm}
\end{equation}
where, $\hat{E}_{lang}$ is the ALN linguistic embeddings after mean pooling over $\hat{e}_{lang}$, i.e. $\hat{E}_{lang}={MeanPool}(\hat{e}_{lang})$.
\vspace{-0.3cm}
\subsection{Cross-attention Layer}

We design the transfer learning layer to map the acoustic embeddings to linguistic embeddings as it is optimized with respect to the BERT embeddings. 
We employ a cross-attention method as in~\cite{xu2019learning},\cite{tsai2019MULT} to generate the alignment between the two streams of embedding vectors, namely, acoustic embeddings, and ALN linguistic embeddings. In traditional attention network, a decoder learns to attend some parts of encoder output~\cite{bahdanau2014neural,chorowski2015attention}. Here, we leverage the attention mechanism to learn the alignment weights between the two embeddings.

Before applying the cross-attention, we use a mapping layer to match the dimension (256-dimensional) of both the embedding sequences. The output of the cross-attention layer can be expressed as
\vspace{-0.2cm}
\begin{equation}
    z = \text{softmax}\left( \frac{{}f_q(X_q)(f_k(X_k))^T}{\sqrt{d_k}}\right)f_v(X_v),
    \vspace{-0.1cm}
\end{equation}
where, $f_q$, $f_k$ and  $f_v$ represent the linear layers for query $X_q$, key $X_k$ and value $X_v$ components, respectively, $d_k$ is the embedding dimension (256), and $z$ is the output from the cross attention layer, which we refer to as {\it ALN embeddings}. We use $e_{acoust}$ as query component, while $\hat{e}_{lang}$ is considered as key and value components.

\vspace{-0.2cm}
\subsection{Intent Layer}\label{Intent_Classification_Layer}
The intent layer aims to classify intents from the ALN embeddings. The first part of this module is a recurrent neural network, which is gated recurrent unit (GRU), followed by maxpooling and linear layer for intent classification. The intent classification loss is refered to as ${Loss}_{intent}$.
\vspace{-0.2cm}
\begin{equation}\label{Intent_loss}
Loss_{intent}=CrossEntropy(f(z),y),
\vspace{-0.2cm}
\end{equation}
where, $f(z)$ is the intent layer output and $y$ represents intent labels. We combine ${Loss}_{intent}$ with the transfer learning loss (${Loss}_{TL}$) using the weight $\alpha$ to derive the total loss, which is used to backpropagate the ALN framework.
\vspace{-0.2cm}
\begin{equation}
    Loss_{{total}} = \alpha Loss_{TL} + (1- \alpha) Loss_{intent}
    \label{eq:totalloss}
    \vspace{-0.2cm}
\end{equation}

\section{Experiments}\label{Experiments}
\subsection{Database}\label{DatabaseSection}
\vspace{-0.2cm}
We evaluate the systems on two databases. The ATIS corpus is one of the most commonly used long-standing datasets for text-based SLU research~\cite{price1990evaluation}. To use the ATIS database for acoustic feature based SLU task, we have rearranged the train and test sets so that corresponding audio files are available from the original ATIS corpus recordings. We have filtered the ATIS database to make sure that all utterances have original speech recordings. The utterances in the Fluent speech commands (FSC) dataset~\cite{lugosch2019speech} serve as speech commands to a virtual assistant. Each command consists of three slot values, namely, action, object and location. The combination of three slot values represents the intent. The details of both the databases is presented in Table~\ref{Database}.

\begin{table}[h]
\vspace{0.3cm} 
\centerline{
{\ninept \caption{\label{Database} {Statistics of ATIS and Fluent speech commands (FSC) databases.}}}}
\renewcommand{\arraystretch}{1.2}
\vspace{-0.1cm}
\resizebox{0.45\textwidth}{!}{
\begin{tabular}{|c c c c c|}
\hline
\multicolumn{1}{|c|}{\multirow{2}{*}{\backslashbox{Specification}{Database} }} & \multicolumn{2}{|c|}{\bf{ATIS}}  & \multicolumn{2}{|c|}{\bf{FSC}}\\
\cline{2-5}
\multicolumn{1}{|c|}{} & \multicolumn{1}{|c|}{Train} & \multicolumn{1}{|c|}{Test} & \multicolumn{1}{|c|}{Train} & \multicolumn{1}{|c|}{Test}\\
\hline
\hline
\multicolumn{1}{|c|}{\#Utterances} & \multicolumn{1}{|c|}{5,253} & \multicolumn{1}{|c|}{580} & \multicolumn{1}{|c|}{23,132} & \multicolumn{1}{|c|}{3,793}\\
\hline
\multicolumn{1}{|c|}{Duration (min)} & \multicolumn{1}{|c|}{569.90} & \multicolumn{1}{|c|}{61.66} & \multicolumn{1}{|c|}{882.95} & \multicolumn{1}{|c|}{154.58}\\
\hline
\multicolumn{1}{|c|}{Avg. utterance len (sec)} & \multicolumn{1}{|c|}{6.51} & \multicolumn{1}{|c|}{6.37} & \multicolumn{1}{|c|}{2.29} & \multicolumn{1}{|c|}{2.45}\\
\hline
\multicolumn{1}{|c|}{\#Intents} & \multicolumn{1}{|c|}{15} & \multicolumn{1}{|c|}{15} & \multicolumn{1}{|c|}{31} & \multicolumn{1}{|c|}{31}\\
\hline
\end{tabular}
}
\vspace{-.4cm}
\end{table}

\subsection{Experimental Setup}\label{Experimentsetup}
\vspace{-0.1cm}
\subsubsection{Baseline systems}
\vspace{-0.2cm}
To show the comparative performance of the proposed ALN framework, we develop two baseline systems. First, the conventional pipeline approach ({\bf Baseline-1}), where we initially pass the speech signal through an ASR system to derive the transcription, which are further applied to the fine-tuned BERT model for intent classification. The ASR system employed here is an in-house general purpose  ASR trained  using  a  combination  of  several  speech  databases. We  haven't  adapted  any  component of  the  ASR  system  specific  to  the  current  application.  The second baseline~\cite{lugosch2019speech} is deployed using acoustic features extracted from the pretrained ASR model ({\bf Baseline-2}) in Section~\ref{Pretrained_ASR_Model}, followed by the intent layer discussed in Section~\ref{Intent_Classification_Layer}. We note that, in this case only intent loss is used as in Equation~\ref{Intent_loss} to optimize the network.
\vspace{-0.2cm}
\subsubsection{Proposed systems}
\vspace{-0.2cm}
In the proposed ALN framework, we use a pretrained ASR model and a finetuned BERT model to extract the acoustic and linguistic embeddings, respectively. The pretrained ASR model used in this work is adopted from~\cite{lugosch2019speech}, which is frozen and not modified for downstream task. For the finetuned BERT model, we use train set of both databases as per Table \ref{Database}. For the two stage-finetuning of BERT model as described in Section \ref{sec:pretrained-nlu}, during MLM stage we use Adam optimizer with learning rate 5e-5, $\beta_1=0.9$, $\beta_2$=0.999 and 3 epochs~\cite{MLM_finetuning}. For intent classification stage of finetuning, we use Adam optimizer with learning rate of 2e-5 and 4 epochs.

To demonstrate the effectiveness of the transfer learning strategy and subsequent improvement with the proposed ALN framework, we develop two systems. In the first system ({\bf ALN linguistic}), initially we pass the acoustic embeddings through the transfer learning layer to obtain the ALN linguistic embeddings.  Then we use only ALN linguistic embeddings in the intent layer described in Section~\ref{Intent_Classification_Layer} for intent classification. In the second system ({\bf ALN}), we obtain the alignment between the ALN linguistic embeddings and acoustic embeddings using the cross attention layer. The output of the cross attention layer is passed through the intent layer to derive the intent classification as shown in Figure~\ref{fig:block-diagram}.  During training of the ALN, the total loss as per Equation~\ref{eq:totalloss} is backpropagated for transfer learning layer, cross-attention layer and intent layer simultaneously.
The ALN linguistic, ALN, and Baseline-2 frameworks are implemented using using PyTorch. We use the Adam optimizer with a learning rate of 0.001, batch size 64 and 100 epochs.

 \vspace{-0.3cm}
\subsection{Results}
 \vspace{-0.1cm}
Table \ref{tab:Baseline} lists the intent classification accuracy of the Baseline-1 and Baseline-2 for both the databases. We also show the pipeline approach when groundtruth text is used instead of ASR output text. We note that WER for ATIS and FSC databases are 11.57\% and 20.43\%, respectively.
The Baseline-2 gives better performance (98.80\%) when the WER is higher as in case of FSC dataset. Whereas, Baseline-1 gives better performance in case of the ATIS database (89.65\%), where WER is lower. However, the performance of the pipeline approach using groundtruth transcription is better than Baseline-2. This indicates the involvement of linguistic information to improve intent classification.

\begin{table}[t]
\centerline{
{\ninept \caption{
\label{tab:Baseline} {Intent classification accuracy of Baseline-1 (pipeline approach) using ASR output and ground truth text, and Baseline-2 (Pretrained ASR) frameworks.}}}
\renewcommand{\arraystretch}{1}
\resizebox{0.45\textwidth}{!}{
\begin{tabular}{|c c c c c|}
\hline
\multicolumn{1}{|c|}{\multirow{2}{*}{\bf Database}} & \multicolumn{3}{|c|}{\bf{Baseline-1}}  & \multicolumn{1}{|c|}{\bf{Baseline-2~\cite{lugosch2019speech}}}\\
\cline{2-4}
\multicolumn{1}{|c|}{} & \multicolumn{1}{|c|}{Ground truth} & \multicolumn{1}{|c|}{ASR output} & \multicolumn{1}{|c|}{WER(\%)} & \multicolumn{1}{|c|}{\bf }\\
\hline
\hline
\multicolumn{1}{|c|}{ATIS} & \multicolumn{1}{|c|}{93.73} & \multicolumn{1}{|c|}{89.65} & \multicolumn{1}{|c|}{11.57} & \multicolumn{1}{|c|}{85.34}\\
\hline
\multicolumn{1}{|c|}{FSC} & \multicolumn{1}{|c|}{100} & \multicolumn{1}{|c|}{95.33} & \multicolumn{1}{|c|}{20.43} & \multicolumn{1}{|c|}{98.80}\\
\hline
\end{tabular}}
}
\vspace{-.5cm}
\end{table}

\begin{figure}[t]
\vspace{-.3cm}
\centering
\centerline{\epsfig{figure=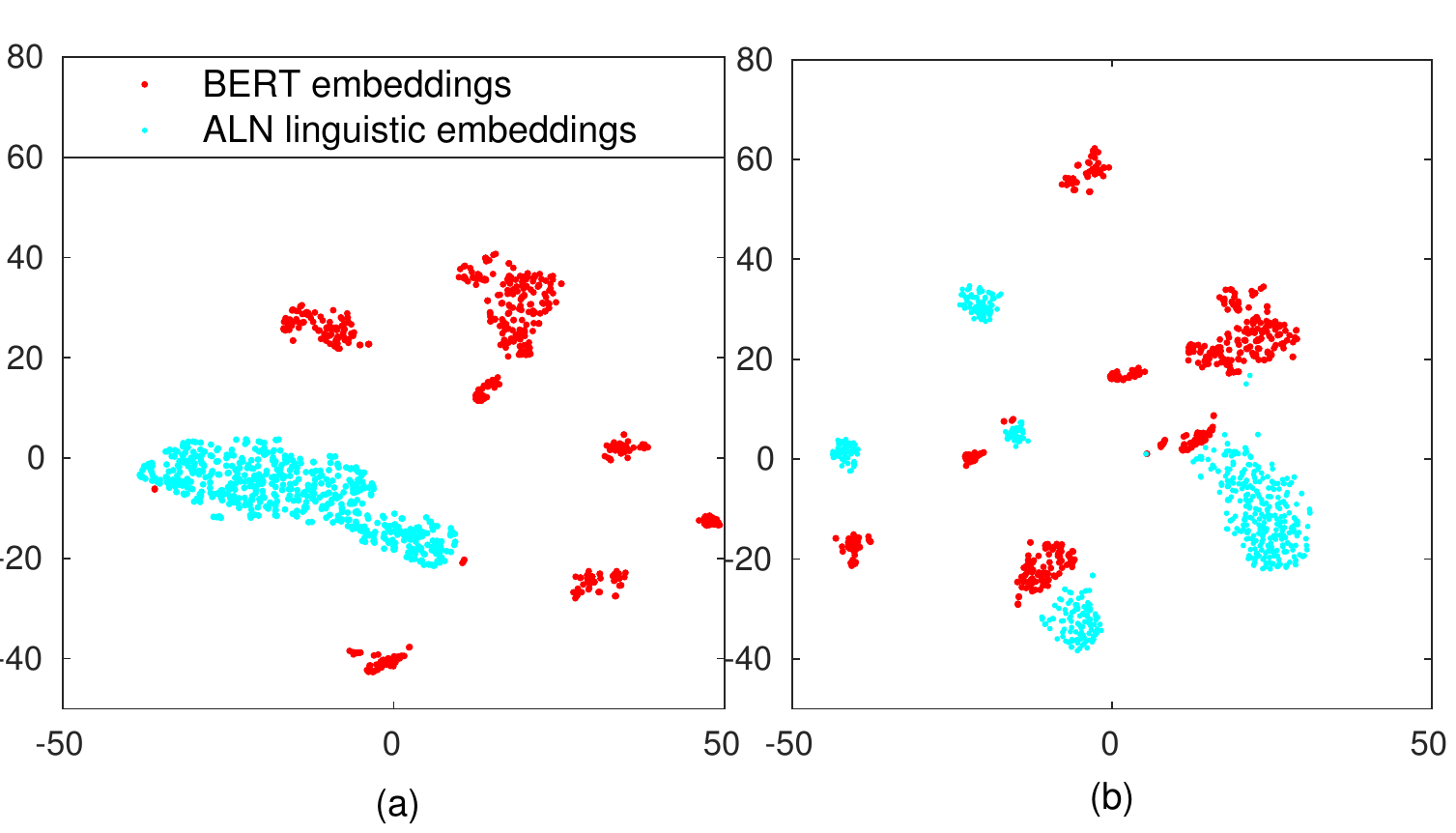, height=1.65in,width=3.2in}}
\vspace{-.4cm}
{\ninept\caption{t-SNE visualization plots to compare BERT embeddings with ALN linguistic embedding in (a) initial epoch, (b) final epoch for 6 intent classes of the ATIS database .}
\vspace{-.4cm}
\label{fig:TSNE}}
\end{figure}

To illustrate the effect of transfer learning in ALN, we visualize t-Distributed Stochastic Neighbor Embedding (t-SNE) representation~\cite{maaten2008visualizing} of embeddings.  Figure~\ref{fig:TSNE}(a) and Figure~\ref{fig:TSNE}(b) show the t-SNE plots across BERT embeddings and ALN linguistic embeddings in an initial epoch and final epoch, respectively, for the ATIS database. We observe that the ALN linguistic embeddings are closer to BERT embeddings in Figure~\ref{fig:TSNE}(b) than in Figure~\ref{fig:TSNE}(a). We note similar embeddings visualization behavior  for FSC database.

In Table \ref{tab:proposed} we report performance of the ALN linguistic and ALN frameworks. We also demonstrate the effect of weight, $\alpha$ in Equation~\ref{eq:totalloss}, with experiments for two values of $\alpha$ (0.5 and 0.8) in total loss computation. In both the frameworks, we observe relatively better performance using $\alpha=0.8$. This indicates that the transfer learning loss contributes more in learning better ALN linguistic embeddings. The intent classification accuracy of ALN linguistic (87.75\%) is better than that of Baseline-2 (85.34\%) for ATIS database. For FSC database the performance of  Baseline-2 (98.80\%) is slightly better that that of ALN linguistic (98.31\%). This shows the efficacy of the ALN linguistic embeddings derived through the transfer learning layer. 

It is evident from Table~\ref{tab:proposed} that the proposed ALN framework outperforms both baseline systems presented in Table \ref{tab:Baseline}. For $\alpha=0.8$ the accuracy of the proposed ALN framework is $90.86\%$ for ATIS database, and  $99.07\%$ for FSC database. The same for Baseline-2 are $85.34\%$ and $98.80\%$, respectively. We observe the performance is consistently higher for both the values of $\alpha$ for ALN framework, compared to Baseline-1 and Baseline-2, for the two databases. For the ATIS database the performance of the ALN framework with $\alpha=0.5$ is lower than that of the Baseline-1. This may be because of the lower WER of the ATIS database and insufficient representation of ALN linguistic features using $\alpha=0.5$.

\begin{table}[ht]
\vspace{0.3cm} 
\centerline{
\ninept \caption{\label{tab:proposed} {Intent classification accuracy (\%) using ALN linguistic and proposed ALN frameworks for different values of weight $\alpha$.}}}
\renewcommand{\arraystretch}{1}
\vspace{-0.2cm}
\scalebox{1}{
\begin{tabular}{|c c c c|}
\hline
\multicolumn{1}{|c|}{\multirow{1}{*}{\bf Database}}  & \multicolumn{1}{|c|}{\bf $\alpha$ value} & \multicolumn{1}{|c|}{\bf ALN linguistic} & \multicolumn{1}{|c|}{\bf ALN}\\
\hline
\hline
\multicolumn{1}{|c|}{\multirow{2}{*}{ATIS}} & \multicolumn{1}{|c|}{0.5} & \multicolumn{1}{|c|}{85.00} & \multicolumn{1}{|c|}{86.37}\\
\cline{2-4}
\multicolumn{1}{|c|}{} & \multicolumn{1}{|c|}{0.8} & \multicolumn{1}{|c|}{87.75} & \multicolumn{1}{|c|}{\bf 90.86}\\
\hline
\multicolumn{1}{|c|}{\multirow{2}{*}{FSC}}  & \multicolumn{1}{|c|}{0.5} & \multicolumn{1}{|c|}{97.20} & \multicolumn{1}{|c|}{99.02}\\
\cline{2-4}
\multicolumn{1}{|c|}{}  & \multicolumn{1}{|c|}{0.8} & \multicolumn{1}{|c|}{98.31} & \multicolumn{1}{|c|}{\bf 99.07}\\
\hline
\end{tabular}
}
\vspace{-.4cm}
\end{table}
\vspace{-0.3cm}
\section{Conclusions}\label{Conclusion}
\vspace{-0.2cm}
This paper presents an end-to-end intent classification framework using both acoustic and linguistic embeddings extracted from a speech signal, without using any intermediate text representation. We extract the acoustic features from a pretrained ASR model and we learn the linguistic features from a pretrained BERT model. We use transfer learning technique to convert acoustic features towards linguistic features. In particular, we employ the teacher-student transfer learning approach to leverage the linguistic information and incorporate into the intent classification network. Finally, we fuse both the acoustic and linguistic embeddings effectively through a cross-attention module. 

 The two modalities, speech and text corresponding to a spoken utterance carry contrasting and crucial behavior to interpret it's meaning. Through the proposed method, we establish the impact of using both  acoustic and linguistic modalities for SLU task. Our experimental results indicate that the transfer learning can learn the information from the linguistic embeddings and perform better than using only acoustic information as well as conventional pipeline SLU approach. 

\newpage
\footnotesize
\bibliographystyle{IEEEbib}
\bibliography{IntentClassification}

\begin{thebibliography}{10}

\bibitem{tur2011spoken}
G{\"o}khan T{\"u}r and Renato De~Mori,
\newblock {\em Spoken language understanding: Systems for extracting semantic
  information from speech},
\newblock John Wiley \& Sons, 2011.

\bibitem{bhargava2013easy}
Aditya Bhargava, Asli Celikyilmaz, Dilek Hakkani-T{\"u}r, and Ruhi Sarikaya,
\newblock ``Easy contextual intent prediction and slot detection,''
\newblock in {\em International conference on acoustics, speech and signal
  processing (ICASSP)}, 2013, pp. 8337--8341.

\bibitem{ravuri2015recurrent}
Suman Ravuri and Andreas Stolcke,
\newblock ``Recurrent neural network and lstm models for lexical utterance
  classification,''
\newblock in {\em INTERSPEECH}, 2015, pp. 135--139.

\bibitem{sarikaya2014application}
Ruhi Sarikaya, Geoffrey~E Hinton, and Anoop Deoras,
\newblock ``Application of deep belief networks for natural language
  understanding,''
\newblock {\em IEEE/ACM Transactions on Audio, Speech, and Language
  Processing}, vol. 22, no. 4, pp. 778--784, 2014.

\bibitem{qian2017exploring}
Yao Qian, Rutuja Ubale, Vikram Ramanaryanan, Patrick Lange, David
  Suendermann-Oeft, Keelan Evanini, and Eugene Tsuprun,
\newblock ``Exploring {ASR}-free end-to-end modeling to improve spoken language
  understanding in a cloud-based dialog system,''
\newblock in {\em Automatic Speech Recognition and Understanding Workshop
  (ASRU)}, 2017, pp. 569--576.

\bibitem{serdyuk2018towards}
Dmitriy Serdyuk, Yongqiang Wang, Christian Fuegen, Anuj Kumar, Baiyang Liu, and
  Yoshua Bengio,
\newblock ``Towards end-to-end spoken language understanding,''
\newblock in {\em International Conference on Acoustics, Speech and Signal
  Processing (ICASSP)}, 2018, pp. 5754--5758.

\bibitem{amodei2016deep}
Dario Amodei, Sundaram Ananthanarayanan, Rishita Anubhai, Jingliang Bai, Eric
  Battenberg, Carl Case, Jared Casper, Bryan Catanzaro, Qiang Cheng, Guoliang
  Chen, et~al.,
\newblock ``Deep speech 2: End-to-end speech recognition in english and
  mandarin,''
\newblock in {\em International conference on machine learning}, 2016, pp.
  173--182.

\bibitem{chan2016listen}
William Chan, Navdeep Jaitly, Quoc Le, and Oriol Vinyals,
\newblock ``Listen, attend and spell: A neural network for large vocabulary
  conversational speech recognition,''
\newblock in {\em International Conference on Acoustics, Speech and Signal
  Processing (ICASSP)}, 2016, pp. 4960--4964.

\bibitem{soltau2016neural}
Hagen Soltau, Hank Liao, and Hasim Sak,
\newblock ``Neural speech recognizer: Acoustic-to-word {LSTM} model for large
  vocabulary speech recognition,''
\newblock in {\em INTERSPEECH}, 2017, pp. 3707--3711.

\bibitem{oord2016wavenet}
A{\"{a}}ron van~den Oord, Sander Dieleman, Heiga Zen, Karen Simonyan, Oriol
  Vinyals, Alex Graves, Nal Kalchbrenner, Andrew~W. Senior, and Koray
  Kavukcuoglu,
\newblock ``Wavenet: A generative model for raw audio,''
\newblock in {\em {ISCA} Speech Synthesis Workshop}, 2016, p. 125.

\bibitem{sutskever2014sequence}
Ilya Sutskever, Oriol Vinyals, and Quoc~V Le,
\newblock ``Sequence to sequence learning with neural networks,''
\newblock in {\em Advances in Neural Information Processing Systems 27}, pp.
  3104--3112. 2014.

\bibitem{bahdanau2014neural}
Dzmitry Bahdanau, Kyunghyun Cho, and Yoshua Bengio,
\newblock ``Neural machine translation by jointly learning to align and
  translate,''
\newblock in {\em International Conference on Learning Representations,
  {ICLR}}, 2015.

\bibitem{gehring2017convolutional}
Jonas Gehring, Michael Auli, David Grangier, Denis Yarats, and Yann~N Dauphin,
\newblock ``Convolutional sequence to sequence learning,''
\newblock in {\em International Conference on Machine Learning, {ICML}}, 2017,
  pp. 1243--1252.

\bibitem{haghani2018audio}
Parisa Haghani, Arun Narayanan, Michiel Bacchiani, Galen Chuang, Neeraj Gaur,
  Pedro Moreno, Rohit Prabhavalkar, Zhongdi Qu, and Austin Waters,
\newblock ``From audio to semantics: Approaches to end-to-end spoken language
  understanding,''
\newblock in {\em IEEE Spoken Language Technology Workshop (SLT)}, 2018, pp.
  720--726.

\bibitem{lugosch2019speech}
Loren Lugosch, Mirco Ravanelli, Patrick Ignoto, Vikrant~Singh Tomar, and Yoshua
  Bengio,
\newblock ``Speech model pre-training for end-to-end spoken language
  understanding,''
\newblock in {\em INTERSPEECH}, 2019, pp. 814--818.

\bibitem{renkens2018capsule}
Vincent Renkens and Hugo~Van hamme,
\newblock ``Capsule networks for low resource spoken language understanding,''
\newblock in {\em INTERSPEECH}, 2018, pp. 601--605.

\bibitem{poncelet2020multitask}
Jakob Poncelet and Hugo~Van hamme,
\newblock ``Multitask learning with capsule networks for speech-to-intent
  applications,''
\newblock in {\em International Conference on Acoustics, Speech and Signal
  Processing (ICASSP)}, 2020, pp. 8494--8498.

\bibitem{mitra2019leveraging}
Vikramjit Mitra, Sue Booker, Erik Marchi, David~Scott Farrar, Ute~Dorothea
  Peitz, Bridget Cheng, Ermine Teves, Anuj Mehta, and Devang Naik,
\newblock ``Leveraging acoustic cues and paralinguistic embeddings to detect
  expression from voice,''
\newblock in {\em INTERSPEECH}, 2019, pp. 1651--1655.

\bibitem{lugosch2019using}
Loren Lugosch, Brett Meyer, Derek Nowrouzezahrai, and Mirco Ravanelli,
\newblock ``Using speech synthesis to train end-to-end spoken language
  understanding models,''
\newblock in {\em International Conference on Acoustics, Speech and Signal
  Processing (ICASSP)}, 2020, pp. 8499--8503.

\bibitem{huang2020leveraging}
Yinghui Huang, Hong-Kwang Kuo, Samuel Thomas, Zvi Kons, Kartik Audhkhasi, Brian
  Kingsbury, Ron Hoory, and Michael Picheny,
\newblock ``Leveraging unpaired text data for training end-to-end
  speech-to-intent systems,''
\newblock in {\em International Conference on Acoustics, Speech and Signal
  Processing (ICASSP)}, 2020, pp. 5754--5758.

\bibitem{denisov2020pretrained}
Pavel Denisov and Ngoc~Thang Vu,
\newblock ``Pretrained semantic speech embeddings for end-to-end spoken
  language understanding via cross-modal teacher-student learning,''
\newblock {\em arXiv preprint arXiv:2007.01836}, 2020.

\bibitem{devlin2019bert}
Jacob Devlin, Ming-Wei Chang, Kenton Lee, and Kristina Toutanova,
\newblock ``{BERT:} pre-training of deep bidirectional transformers for
  language understanding,''
\newblock in {\em North American Chapter of the Association for Computational
  Linguistics: Human Language Technologies, {NAACL-HLT}}, p. 4171–4186.

\bibitem{sentencebert}
Nils Reimers and Iryna Gurevych,
\newblock ``Sentence-{BERT}: Sentence embeddings using {S}iamese
  {BERT}-networks,''
\newblock in {\em Empirical Methods in Natural Language Processing and
  International Joint Conference on Natural Language Processing,
  {(EMNLP-IJCNLP)}}. 2019, pp. 3982--3992, Association for Computational
  Linguistics.

\bibitem{hinton2015distilling}
Geoffrey Hinton, Oriol Vinyals, and Jeff Dean,
\newblock ``Distilling the knowledge in a neural network,''
\newblock {\em arXiv preprint arXiv:1503.02531}, 2015.

\bibitem{li2017large}
Jinyu Li, Michael~L Seltzer, Xi~Wang, Rui Zhao, and Yifan Gong,
\newblock ``Large-scale domain adaptation via teacher-student learning,''
\newblock in {\em INTERSPEECH}, 2017, pp. 2386--2390.

\bibitem{xu2019learning}
Haiyang Xu, Hui Zhang, Kun Han, Yun Wang, Yiping Peng, and Xiangang Li,
\newblock ``Learning alignment for multimodal emotion recognition from
  speech,''
\newblock in {\em INTERSPEECH}, pp. 3569--3573.

\bibitem{tsai2019MULT}
Yao-Hung~Hubert Tsai, Shaojie Bai, Paul~Pu Liang, J.~Zico Kolter,
  Louis-Philippe Morency, and Ruslan Salakhutdinov,
\newblock ``Multimodal transformer for unaligned multimodal language
  sequences,''
\newblock in {\em Annual Meeting of the Association for Computational
  Linguistics (Volume 1: Long Papers)}, 2019.

\bibitem{chorowski2015attention}
Jan~K Chorowski, Dzmitry Bahdanau, Dmitriy Serdyuk, Kyunghyun Cho, and Yoshua
  Bengio,
\newblock ``Attention-based models for speech recognition,''
\newblock in {\em Advances in neural information processing systems}, 2015, pp.
  577--585.

\bibitem{price1990evaluation}
Patti Price,
\newblock ``Evaluation of spoken language systems: The {ATIS} domain,''
\newblock in {\em Speech and Natural Language: Proceedings of a Workshop Held
  at Hidden Valley, Pennsylvania, June 24-27, 1990}, 1990.

\bibitem{MLM_finetuning}
``huggingface/transformers,''
  \url{https://github.com/huggingface/transformers/blob/master/examples/language-modeling/run_language_modeling.py},
\newblock [Online; accessed 21-October-2020].

\bibitem{maaten2008visualizing}
Laurens Maaten and Geoffrey Hinton,
\newblock ``Visualizing data using {t-SNE},''
\newblock {\em Journal of machine learning research}, vol. 9, no. Nov, pp.
  2579--2605, 2008.

\end{thebibliography}

\end{document}